# Semi-supervised Bootstrapping approach for Named Entity Recognition

S. Thenmalar, J. Balaji and T.V. Geetha

Department of Computer Science and Engineering,

Anna University, Chennai.

*Abstract*

  The aim of Named Entity Recognition (NER) is to identify references of named entities in unstructured documents, and to classify them into pre-defined semantic categories. NER often aids from added background knowledge in the form of gazetteers. However using such a collection does not deal with name variants and cannot resolve ambiguities associated in identifying the entities in context and associating them with predefined categories. We present a semi-supervised NER approach that starts with identifying named entities with a small set of training data. Using the identified named entities, the word and the context features are used to define the pattern. This pattern of each named entity category is used as a seed pattern to identify the named entities in the test set. Pattern scoring and tuple value score enables the generation of the new patterns to identify the named entity categories. We have evaluated the proposed system for English language with the dataset of tagged (IEER) and untagged (CoNLL 2003) named entity corpus and for Tamil language with the documents from the FIRE corpus and yield an average f-measure of 75% for both the languages.

## 1. Introduction

In general, proper nouns are considered as named entities. The NER task was introduced during the 6$^{th}$ Message Understanding Conference (MUC) in 1996 [13], and in MUC- 7 [3] the initial classification of named entities used the following categories and subcategories: Entity (ENAMEX): person, organization, location, Time expression (TIMEX): date, time and Numeric expression (NUMEX): money, percent. However, named entity tasks often include as named entities expressions for date and time, names of sports and adventure activities, terms for biological species and substances. The major challenge of named entity recognition is that of tagging sequences of words that represent interesting entities, such as people, places, and organizations. NER is a two-step process, the first step being the identification of proper nouns which is the marking of the presence of a word or phrase as named entity (NE) in a given sentence while the second step is its classification where the role of the identified NE is determined.

NER was initially known as a significant component for Information Extraction (IE). NER has now become vital for many other natural language processing based applications. The identification and the semantic categories of Named entities are necessary before recognizing relations between these entities [10, 11, 36]. NE's play an important role in identifying ontological concepts for populating ontologies [5, 12]. NEs convey the crucial information that drives Information Retrieval (IR) and Question Answering (QA) systems [23, 35]. In more recent times, important applications like news aggregation are usually centred on entities.

There are several approaches for identifying NER. The rule-based approach uses a set of rules defined by human experts to extract entities. This model takes a set of patterns consisting of grammatical, syntactic and orthographic features in combination with dictionaries. However, manual creation of rules is labour intensive and costly and requires significant language as well as domain expertise [33]. Moreover systems developed for one domain cannot be ported to another domain.

Therefore learning based approaches have been introduced for NER. Learning algorithms can be defined as methods that use the features of training data and automatically induce patterns for recognising similar information from unseen data. Learning algorithms can be generally classified into three types: supervised learning, semi-supervised learning and unsupervised learning. Supervised learning utilises only the labelled data to generate a model. Semi-supervised learning aims to combine both the labelled data as well unlabelled data in learning. Unsupervised learning aims to learn without any labelled data.

We present a semi-supervised pattern based bootstrapping approach to NER that automatically identifies and classifies the entities. Our approach starts with the small set of tagged training data. The tagged training data is used to identify the word and context features to define a five window context pattern for each named entity category. We explore the representation of features used for both English and Tamil languages to define the pattern. The identified patterns are used as seed patterns. These seed patterns are used to identify the entities as an exact match in the test set. The pattern scoring and the tuple value scoring decide the modification needed to generate new patterns. The pattern score identifies which set of patterns are used for the next iteration. The tuple value scoring of POS provides which set of tuple contributes to the named entity and decides the window movement that is shift to the left or to the right and masks one tuple thus generating of new patterns that is used to learn new context to identify Named entities.

The rest of the paper is organized as follows. Section 2 gives a brief description of related work especially in the area of machine learning approaches to NER. Section 3 deals with the bootstrapping approach to NER. Section 4 deals with evaluation and results while section 5 gives a conclusion and discusses future work.

## 2. Related Work

In a supervised learning approach an NER system takes training data and their features as input to generate an extraction model, which is then used to identify similar objects in new data. Supervised learning has been the most commonly used and the leading approach in the NER [28]. There are several widely used machine learning techniques for this task. Support Vector Machines (SVM) [15, 7] a model is constructed that fits a hyperplane that best splits positive and negative examples in the labelled data. The model characterizes the examples as points in space, represented so that the positive and negative examples are separated by a clear gap that is as wide as possible. New examples are represented into that same space during the application time and expected to belong to a category based on which side of the gap they fall on.

Hidden Markov Model (HMM) [46, 44, 30] is a statistical Markov model in which the sequence of states are hidden but can be predicted from a sequence of observations conveyed as a probabilistic function of the states. The learning process in the context of NER concludes that an HMM is based on the observed features and tags present in the training data. The model generates a mapping with certain probability that can predict a sequence of states. Conditional Random Fields (CRF) [17, 1] is a probabilistic model which avoids certain assumptions about the input and output sequence distributions of HMM. The other broadly used machine learning techniques for detecting NER where Perceptron algorithms [18], Naïve Bayes [27], Decision Trees [9] and Maximum Entropy model [2]. For Tamil language named entities are identified by using Expectation Maximisation and the CRF model [29, 41]. The identification for named entity using CRF for Tamil language [41] describes the characteristics feature and handles morphological inflections to represent the training model. In our approach, we consider the morphological suffices as an added feature to represent the pattern for the named entity categories. The supervised learning method depends on the large set of training data, which has to be annotated manually.

Unsupervised learning methods recognize named entity based on unlabelled data. The unsupervised learning methods basically use clustering techniques, distribution statistics and similarity based functions. The recognition of various types of named entities in the open domain that could be useful in IE [8]. The sequence of capitalised words that are likely to be named entities are extracted and the search queries using the sequence of words are created with Hearst patterns [14]. The hypernyms were extracted and clustered. The entities are looked up in WordNet and are labelled by the top level concepts observed in the WordNet. The named entities are lexicalised as multi-word in which co-occuring terms occur more frequently [6]. They have identified possible n-grams entity from the corpus based on the mutual information measures and the frequency of words and grouped similar named entities using clustering algorithm. The complex named entities in the Web data is identified using clustering algorithm [45]. The named entities are labelled based on the similarity using vector similarity model [4]. In essence, the entity names and their types are described as vectors with the specified features. In Semantic Concept Mapping, with the known list of candidate entity names and labels are denoted as WordNet synsets [19]. The Lin's similarity function describes the type of entity name [25].

We explore the semi-supervised learning method for NER. The semi-supervised method uses the small number of labelled data to learn and tag a large set of unlabelled data. The self-training algorithm is used for detecting named entity and voted co-training algorithm is used for classifying the named entity [21]. The self-training algorithm that selects the unlabelled instances for the Naïve Bayes classifier [43]. The gene name is recognized by bootstrapping approach [42]. The abstract and the list of genes are available for the set of articles. The gene mentions in the abstract were annotated with the list of gene available for the article. The HMM-based tagger is trained with the annotated data. The lists of entities stated in the document are not usually available in general named entity recognition. In another approach the named entities having Heidelberg Named Entity Resource are classified based on the Wikipedia category using bootstrapping [20]. The inconsistency in classification might be undetermined by placing the articles to the specified category. The features of POS and syntactic structure of the document are used for the semi-supervised learning algorithm; a Self-Training algorithm is used to recognize the named entity [34]. The features used determine the entity boundary and pattern extraction. However in our approach we use the POS feature and the context information of the word to represent the pattern. The CRF with feature induction is used for detecting Hindi NER [24]. The features induction includes word based features, character n-grams, word prefix and suffix and 24 gazetteers. However we explore detecting named entity without gazetteers.

NER problem is considered to be solved for languages such as English but still remains a challenge for resource scarce languages. Moreover NER of informal texts such as tweets and blogs still remains a challenging problem [22]. The issues in handling NER tasks for Indian languages have been described in the survey of Named Entity Recognition [16]. The issues discussed are the major feature commonly followed by NER systems is the capitalization of words. However, the capitalization of word for the named entities is not represented in the Indian languages. The gazetteers of named entities are unavailable for Indian languages. Additionally, spelling variations are common in Indian languages. For example in Tamil language: ராசா (Rasa) - ராஜா (Raja): Person named Raja, புதுசேரி (puducherri) - புதுச்சேரி (pudhuccherri) : Place named Puducherry. The lack of labelled data is the added issue to resource scarce and morphological rich languages. In this work, we use pattern representation of bootstrapping approach which requires only a small set of feature tagged seed samples to learn the context and the features that detect the named entity and its category.

## 3. Bootstrapping process for Named Entity Recognition

Figure 1 shows the overall bootstrapping process for NER. The proposed approach starts with the small set of training examples. The training set of documents is manually annotated with the named entity categories. The annotated training set is pre-processed by identifying the features of the word. We make use of the context of the word to define the pattern. The patterns associated with each category of named entity are identified and used as seed patterns. The test data is processed by matching the features of the word with the pattern. If exact match occurs then the named entity category is identified. Up to this point we have identified and categorized named entities that have features exactly similar to the seed set initially given. However we need to generate new patterns by learning new contexts where these named entities can occur. For this purpose the patterns are scored to identify which patterns can be used for further iteration. The new patterns have also been designed to identify named entity chunks even though the initial seed patterns are associated with only a single word. The new pattern is generated by right or left shift of the window for each pattern depending on the tuple value score. The new pattern is given as input to the tagged test documents and the instances of the named entity categories are identified. The process of pattern scoring, tuple value scoring and window shifting to learn new patterns is continued until no new patterns are generated or all the test data is labelled. The feature set used is described in the next section.

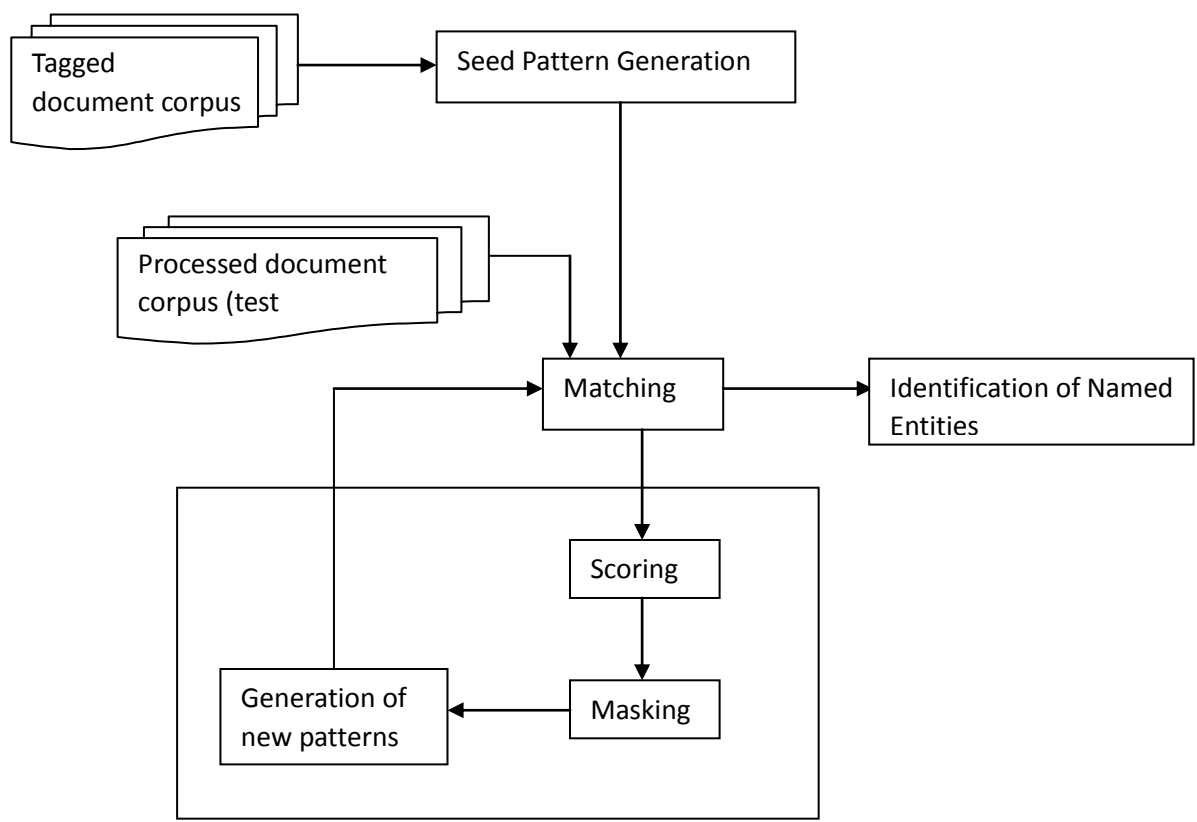

Figure 1 Overall bootstrapping process for Named Entity Recognition

## 3.1 Feature set

A perceptron based recognizer for identifying named entities uses nonlocal dependencies and external information as features [31]. A supervised learning method with CRF for detecting NER uses local knowledge features, external knowledge features and the non-local dependencies [38]. They have discussed that the system when using the local knowledge feature performs poorly when using the single token and the maximum observation of named entities is shown when using the sliding window of 3 token. However, when considering a three window context, the ambiguity of the type of named entity occurs. The features used also depend on the language under consideration. In order to overcome the above issue we go for a five window context *($w_{i-2}, w_{i-1}, w_i, w_{i+1}, w_{i+2}$)*

The commonly used features for NER systems were part-of speech tags, shallow parsing and the gazetteers. Part-of-speech (POS) tags are commonly used feature in NER but this feature is not considered for NER Systems [31, 26]. In this work, we use POS tag and semantic constraints obtained from UNL KB [40] that are associated with each word along with the five window context as common feature for both English and Tamil languages. Thus each word is attached with POS tag and semantic constraint to form the feature set. Using these features we describe the patterns.

### 3.2 Seed pattern Generation

Tagging the words in the English documents with POS tag is carried out using the Stanford parser and the POS tagging of the words in the Tamil documents is carried out using morphological analyser. We manually label the named entities in a small set of tagged data. This data is considered as training data and is used for identifying the patterns. The patterns capture both the sequence of tokens that identify a potential named entity and the information from the right & left context where it occurs. We consider the five window context $(w_{i-2}, w_{i-1}, w_i, w_{i+1}, w_{i+2})$ to represent the pattern. We have defined two types of patterns, one type for English language where the POS tag and semantic constraint (SC) are the features associated with each word is given below. In the case the pattern type of Tamil language we also use morphological suffix (MS) which implicitly conveys case information for nouns as an additional feature is given below. This difference in the type of patterns essentially caters to two languages having different characteristics such as fixed word order of English language and partially free word order of Tamil language. The use of word based semantic constraint (SC) allows the context of the named entity is to be semantically described to enable proper classification. Thus each word in the context window $(w_{i-2}, w_{i-1}, w_i, w_{i+1}, w_{i+2})$ consists two tuples in the case of English language and three tuples in the case of Tamil language.

**Pattern Type for English language**

POS,SC($w_{i-2}$);POS,SC($w_{i-1}$);POS($w_i$),SC@Named Entity Type; POS,SC($w_{i+1}$);POS,SC($w_{i+2}$)

**Pattern Type for Tamil language**

POS,MS& SC($w_{i-2}$);POS,MS&SC($w_{i-1}$);POS,MS & SC ($w_i$) @Named Entity Type; POS,MS&SC($w_{i+1}$);POS,MS&SC($w_{i+2}$)

The Named Entity Types used here are defined in the the classification of MUC-7 namely person, organization, location, date, time, money and percentage.

**Example Pattern Types for English language**

Person
VBG,icl>person; IN,aoj>thing; NNP,iof>person@person; IN,aoj>thing; PRP,icl>female person

Location
TO,aoj>thing; VB,agt>thing,obj>thing; NP,iof>country@location; NN,icl>area; NNS,icl>action

Organization
IN,obj>thing; IN,aoj>thing; nnp,icl>organization@organization; JJ,aoj>thing; NN,icl>facilities

Date
NN,icl>organization; IN,aoj>thing; NNP,icl<date; CD,None@date; VBD,obj>thing; NNS,aoj>thing

Time
NN,icl>action; VBD,None; NN,icl>time@time; IN,aoj>thing; DT,None

Money

NN,icl>deal; IN,aoj>thing; $CD,icl>money@money; IN,aoj>thing; NN,icl>reduce

Percent

VB;agt>thing,obj>thing; PRP,None; CD,None;NN,icl>ratio@percent; NN,agt>thing; IN,aoj>thing

**Example Pattern Type for Tamil language**

Person

Entity,அ,icl>region; Noun,None,icl>person; Entity,None,iof>person@person; Adjective,ப்,icl>help; Adjective,None,aoj>thing

Location

Noun,இன்,icl<weather; Noun,ஆக,iclplace@location; Noun,இல்,icl<area; Noun,None,icl>calculate(agt>thing,obj>thing)

Organization

Noun,None,icl>person; Noun,உடன்,icl>act; Noun,None,icl>organization@organization; Noun,கள்,icl>person; Verb,None,icl>action

Date

Adjective,None,mod<thing; DateTime,None,icl>period; DateTime,None,icl>month, charNumbers,ஆம்@date; DateTime,None,aoj>thing; Noun,உக்கு+ச்,icl>facilities

Money

Pronoun,None,icl>person; Noun,None,aoj>thing; charNumbers,None,Noun,ஐ, icl>currency@money;Adverb,None,icl>action; Noun,ஆர்,aoj>thing

Time

Noun,None,icl>morning; charNumbers,None,Noun,க்கு,icl>time; Noun,None, icl>workship; Verb,உம்,icl>action

Percent

Noun,கள்,icl>person; Noun,க் + கையில்,icl>action; charNumbers,None,Noun, icl>ratio @percent; Noun,ஆக,icl>change; Verb,None,agt>thing,gol>person,obj>thing

### 3.3 Matching

For a given test data we POS tag the words using Stanford parser in case of English language and use a morphological analyser [39] in the case of Tamil language for POS and Morphological suffix tagging. The root words are then used to obtain the corresponding semantic constraints from the UNL KB [40]. We check for the matching of seed patterns with the annotated sentences of the documents. Although the pattern consists of a five word window ($w_{i-2}, w_{i-1}, w_i, w_{i+1}, w_{i+2}$), actual exact matching is carried out with only the middle word $w_i$ of the pattern. If there is a match the corresponding classes are labelled for the exactly matched patterns. Named entities that are not handled by the exact match are processed through partial matching. Partial match is carried out after selecting the pattern to be modified during iteration in the next cycle. Once a pattern is identified, tuple scoring detects which tuple contributes the most to the particular entity type.

### 3.4 Generation of new patterns

The first step in new pattern generation is to find the most frequently occurring pattern for each class of named entity indicated by Pattern Score. The next step is to find alternate values for POS tags that can occur at position $k$ in the context window of pattern $P_j$ keeping all other tuple values the same by finding the tuple value of POS tag with minimum score at position $k$ which is then considered for masking.

#### 3.4.1 Pattern Score

In this work we use the Basilisk algorithm for calculating the pattern scoring metric RlogF metric [32]. The extraction pattern is scored using the following formula:

$$Ps(P_j) = \frac{F_j}{n_j} * \log(F_j) \tag{1}$$

Where $F_j$ is the number of identified named entities by pattern $P_j$ corresponding to a particular type of named entity, $n_j$ is the total number of patterns identified. The pattern score identifies the pattern for the particular type of named entity to be chosen for modification to form the new pattern.

#### 3.4.2 Tuple value Score

The tuple scoring basically depends on the tuple corresponding to the POS tag of the words in the context window. This scoring essentially evaluates which POS tag value of which word in the context window of the specific pattern is strongly associated with a particular pattern.

Let us consider a pattern $P_j$ corresponding to a particular type of Named entity. The pattern $P$ associated with Named entity has four words $(w_{i-2}, w_{i-1}, w_i, w_{i+1}, w_{i+2})$ in the 5 word context window and each of these words is associated with a POS values $(pos_{i-2}, pos_{i-1}, pos_{i+1}, pos_{i+2})$. The maximum score of the POS value indicates that this POS value in this position contributes the most to the pattern $P_j$ and is given as

$$score(tv_{pos_k}) = \frac{\log(f(tv_{pos_k}, P_j))}{f(P_j)*f(tv_{pos_k})} \quad \text{where k=i-2,i-1,i+1,i+2} \tag{2}$$

Here, $tv_{POSk}$ corresponds to tuple value of POS tuple $pos_k$.
$f(tv_{POSk}, P_j)$ is the number of times this particular POS value at position $k$ occurs with pattern $P_j$.
$f(P_j)$ is the frequency of pattern $P_j$ and
$f(tv_{POSk})$ is the total frequency of this tuple value at position $k$.

#### 3.4.3 Methods for New Pattern Generation

The first method of new pattern generation is the replacement of POS value in the appropriate position $k$. The POS tuple value with minimum score at position k is masked. The new pattern is generated by replacing the tuple value of the original pattern by the new tuple value that occurs the most frequently at position $k$ in the test data.

The next method of new pattern generation is carried out by shifting the context window to the left or right of the Wi depending on the frequency of occurrence of POS pair of $w_i$, $w_{i+1}$ or $w_{i-1}$, $w_i$. Depending on a higher POS pair frequency score a new 5 word pattern is generated where either $w_{i-1}$ or $w_{i+1}$ becomes the new $w_i$. This method of new pattern generation is possible since Named Entities are often associated with POS tags that frequently tend to co-occur together.

New patterns are also generated by chunking words to form phrasal named entities. For this purpose again we use the POS pair frequency score as shown in Eq. 1, but in addition we check whether each of the words associated with POS pair have the same semantic constraints. In case the POS pair frequency is above a threshold and have same semantic constraints they are chunked as a single Named entity and considered as $w_i$ for the next iteration. This unique way of forming patterns for chunked words forming Named Entities is possible because these chunks are often associated with similar semantic properties. However the two languages we considered needed to be tackled differently during the chunking process.

$$POS pair frequency\ score = \frac{f(POS(W_i), POS(X))}{f(POS(W_i))} \quad (3)$$

In the case of English language, the POS pair frequency score and semantic constraints alone decide chunking. However in the case of Tamil language, in addition to the above features, morphological suffix should not be associated with Wi in case of pair Wi, Wi+1 or Wi-1 in case pair Wi-1, Wi. In case such morphological suffixes exist chunking is not carried out even though the semantic constraints between the pair matches.

## 4. Evaluation

We have tested the performance of the system on IEER dataset, the tagged corpus which contains the Newswire development test data for the NIST 1999 IE-ER Evaluation. There were totally 3174 named entities. We have taken two different seed patterns that commonly occur for each named entity class (MUC-7). Iterations are carried out until no changes occur in the patterns. The performance of the system is evaluated using precision, recall, and f-measure metrics. Here, recall is defined as

$$recall = \frac{number\ of\ correctly\ recognized\ NEs}{Total\ number\ of\ NEs} \quad (4)$$

Precision is defined as

$$precision = \frac{number\ of\ correctly\ recognized\ NEs}{Total\ number\ of\ NEs\ retrieved\ by\ the\ system} \quad (5)$$

F-measure is the weighted harmonic mean of precision and recall and is given as

$$F - measure = \frac{2 * p * r}{p + r} \quad (6)$$

The performance of NER system using IEER data set is given in Table 1, where the system identifies the named entities with the average precision of 83% and recall of 92%.

| Named Entities | precision | recall | F-measure |
|---|---|---|---|
| person | 84.72 | 93.29 | 88.8 |
| Location | 82.1 | 92.9 | 87.17 |
| organization | 83.08 | 92.19 | 87.4 |
| Date | 83.3 | 92.83 | 87.81 |
| Time | 81.53 | 91.37 | 86.17 |
| Money | 81.55 | 91.3 | 86.15 |
| Percent | 81.48 | 91.67 | 86.28 |

**Table 1 Performance of Bootstrapping system using IEER dataset**

We have tested the performance of the system by using the CoNLL 2003 [37] data RCV1 (Reuters Corpus Volume 1). The Reuters corpus consists of news articles between August 1996 and August 1997. The training set was taken from the files representing the end of august 1996. For test set the files were from December 1996. There were totally 5648 Named Entities. The training data is processed and tagged with POS tags and semantic constraints. We extract two different seed patterns for each named entity class (MUC-7) from the training set. With the seed patterns, we used the test set to identify exact match and iterate the process to generate new patterns for the named entity class. The performance is shown in Table 2, where the system yields the average precision of 80% and recall of 89%.

| Named Entities | precision | recall | F-measure |
|---|---|---|---|
| person | 82.82 | 93.39 | 87.79 |
| location | 81.62 | 90.52 | 85.84 |
| organization | 82.4 | 89.1 | 85.62 |
| date | 79.54 | 87.5 | 83.33 |
| time | 75.84 | 89.5 | 82.11 |
| money | 71.05 | 81 | 75.7 |
| percent | 72.8 | 83 | 77.57 |

**Table 2 Performance of our system for CoNLL 2003**

For Tamil language we performed experiments on documents of the FIRE (Forum of Information Retrieval Evaluation) Tamil corpus extracted from newspapers such as BBC, Dinamani and Dinamalar. We have considered 50000 documents and tagged with the appropriate features such as POS, Morphological suffix and UNL Semantic constraint. We have taken 4000 tagged documents for training set and extracted the most frequently occurring two different example patterns for each named entity class (MUC-7). The performance is shown in Table 3, where the system produces the average precision of 79% and recall of 88%.

| Named Entities | precision | recall | F-measure |
|---|---|---|---|
| person | 84.34 | 90.46 | 87.29 |
| location | 75.16 | 90.01 | 81.92 |
| organization | 77.62 | 87.31 | 82.18 |
| date | 75.15 | 86.3 | 80.34 |
| time | 72.02 | 86.56 | 78.62 |
| money | 73.02 | 86.42 | 79.16 |
| percent | 74.42 | 85.88 | 79.74 |

**Table 3 Performance of our system in FIRE corpus.**

We also compared our bootstrapping with the baseline approach [34]. The Baseline system uses Reuters corpus and considers accident documents. The total number of named entities corresponding to the date and location are 246 and 596. The Baseline system uses 15 and 36 seed patterns for date and location entities and extracted 18 and 68 patterns. Our system uses 2 seed patterns for each named entities and we have learned 12 and 35 patterns of those named entities. The comparison is shown in figure 2, the missing of syntactic structure information in the baseline system yields less F-measure whereas our approach makes use of the larger contextual window and word based semantic information to learn the patterns.

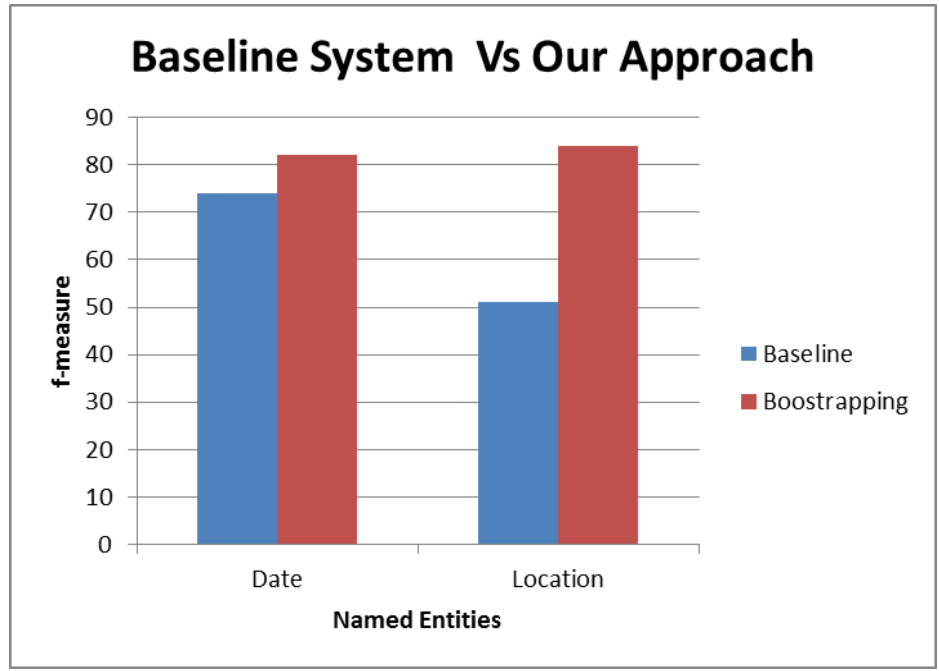

**Figure 2. Performance of Baseline system and our Bootstrapping approach**

We also shown in the Table 4 and Table 5 the number of patterns that we have learnt in the English corpus and Tamil corpus for each named entities using each method (replacement of POS and shifting window).

|  | Number of New Patterns Learnt - CoNLL 2003 | | | | | | |
|---|---|---|---|---|---|---|---|
|  | Person | Location | Organization | Date | Time | Money | Percent |
| Replacement of POS | 26 | 32 | 28 | 17 | 15 | 12 | 8 |
| Shifting the window | 38 | 44 | 35 | 24 | 22 | 18 | 14 |

**Table 4 Number of patterns learnt from the CoNLL 2003 English Corpus**

|  | Number of New Patterns Learnt - FIRE | | | | | | |
|---|---|---|---|---|---|---|---|
|  | Person | Location | Organization | Date | Time | Money | Percent |
| Replacement of POS | 54 | 62 | 58 | 35 | 38 | 18 | 12 |
| Shifting the window | 75 | 83 | 77 | 48 | 42 | 24 | 22 |

**Table 5 Number of patterns learnt from the FIRE Tamil Corpus**

## 5. Conclusion

This paper describes a new pattern based semi-supervised bootstrapping for identifying and classifying Named Entities. The method does not use any Gazetteer but instead uses POS information and word based semantic constraints and gives an average f-measure of 75 % for both the languages. This essentially ensures that the patterns are feature based enabling tagging of hitherto unseen Named Entities. This method can be further enhanced by considering more domain specific corpora and by trying for domain specific Named Entity categories. Future work includes testing this method for other languages to study how the methodology needs to adapted. Pattern definition and chunking strategies are possible aspects that need to be modified.